\documentclass[conference]{IEEEtran}
\IEEEoverridecommandlockouts
\usepackage{cite}
\usepackage{amsmath,amssymb,amsfonts}
\usepackage{algorithm}
\usepackage{algpseudocode}

\usepackage{booktabs}
\usepackage{hhline}
\usepackage{subcaption}  
\usepackage{graphicx}
\usepackage{textcomp}

\usepackage{xcolor}
\usepackage{hyperref}
\usepackage{multirow}
\usepackage{diagbox}
\usepackage{colortbl}
\usepackage[top=0.75in, bottom=1.02in, left=0.58in, right=0.58in]{geometry}
\usepackage{url} 
\def\BibTeX{{\rm B\kern-.05em{\sc i\kern-.025em b}\kern-.08em
    T\kern-.1667em\lower.7ex\hbox{E}\kern-.125emX}}
\begin{document}

\title{Reliable Vertical Federated Learning in 5G Core Network Architecture\\

}

\author{\IEEEauthorblockN{Mohamad Mestoukirdi}
\IEEEauthorblockA{\textnormal{Mitsubishi Electric R\&D Centre Europe
}\\\textnormal{Rennes, France}\\\textnormal{M.Mestoukirdi@fr.merce.mee.com}}
\and
\IEEEauthorblockN{ Mourad Khanfouci}
\IEEEauthorblockA{\textnormal{Mitsubishi Electric R\&D Centre Europe
}\\\textnormal{Rennes, France}\\\textnormal{M.Khanfouci@fr.merce.mee.com}}
}

\maketitle

\begin{abstract}
This work proposes a new algorithm to mitigate model generalization loss in Vertical Federated Learning (VFL) operating under client reliability constraints within 5G Core Networks (CNs). Recently studied and endorsed by 3GPP, VFL enables collaborative and load-balanced model training and inference across the CN. However, the  performance of VFL significantly degrades when the Network Data Analytics Functions (NWDAFs) – which serve as primary clients for VFL model training and inference – experience reliability issues stemming from resource constraints and operational overhead. Unlike edge environments, CN environments adopt fundamentally different data management strategies, characterized by more centralized data orchestration capabilities. This presents opportunities to implement better distributed solutions that take full advantage of the CN data handling flexibility. Leveraging this flexibility, we propose a method that optimizes the vertical feature split among clients while centrally defining their local models based on reliability metrics.  Our empirical evaluation demonstrates the effectiveness of our proposed algorithm, showing improved performance over traditional baseline methods. 
\end{abstract}


\section{Introduction}
In the realm of 5G  networks, there is a growing emphasis on integrating machine learning frameworks into daily operations for both edge devices and operators, driven by the proven advantages these frameworks can offer over traditional model-based methods. Among these, Federated Learning (FL) \cite{b0}  has gained prominence due to its distributed and privacy-preserving framework during both training and inference. FL enables iterative model training across datasets held by distributed clients without compromising privacy.

The modus operandi of FL is contingent upon how data is distributed among the FL clients. Horizontal Federated Learning (HFL) \cite{b0}  focuses on scenarios where a central server, typically without its own dataset, orchestrates the iterative training of a global model. This process involves FL clients with datasets containing different samples, yet derived from the same feature space. In contrast, Vertical Federated Learning (VFL) \cite{VFL} supports model training when different features of available samples are collected by different clients, with labels housed on the server. This expands the server’s role to include loss calculation and gradient back-propagation.

While HFL has been widely studied and deployed across various domains, including 5G standardization, VFL remains less explored, primarily due to its complex requirements for coordinated inference across distributed participants. For instance, HFL has been investigated as a study item by 3GPP and incorporated into Release 18 \cite{b5}, enabling inter- and intra-domain collaborative model training and inference among various Network Data Analytics Functions (NWDAFs). Until recently, VFL has been introduced as a study item in 3GPP \cite{VFLstudy}, with its specifications currently under development in the normative work of SA2. In the literature, numerous studies have tackled the challenges associated with deploying HFL in wireless networks. These challenges include, but are not limited to, the impact of data \cite{ditto} and client heterogeneity \cite{prox} on the model convergence rate, as well as the effects of wireless channel perturbations  on the training process\cite{gunduz}. In contrast, VFL has received less attention, as its unique challenges and deployment scenarios are generally more complex. Previous works mainly focused on features alignment from the point of view of preserving users data privacy\cite{zhang2024treecssefficientframeworkvertical}, decentralized serverless implementation to mitigate server unreliability \cite{unreliableserver}, and reliability of clients\cite{dropout}.

In this work, we explore the challenge of client reliability in VFL within a Core Network (CN) setup, with a particular emphasis on 5G CNs. Central to these networks are the NWDAFs, which, through their Model Training Logical Function (MTLF) and Analytics Logical Function (AnLF), utilize computational hardware to perform model training, inference, and analytics processing respectively. Despite their substantial computing capabilities, these CN functions are expected to serve as central AI enablers for numerous applications, many with demanding performance requirements. This extensive workload could generate significant overhead, potentially compromising their reliability. Accordingly, We propose a solution to mitigate the VFL trained model generalization performance loss encountered in such scenarios through optimized feature distribution and model definition. We then analyze the performance of VFL model training and inference performance under the reliability and availability constraints of these entities under such architectural setups. To the best of our knowledge, this is the first work that addresses client unreliability during training and inference in a VFL setting through the lens of an optimized feature distribution and model definition algorithm.

\section{Problem Formulation}
In a prototypical VFL system with $K$ \textit{passive} clients and an \textit{active} client (aka. server), in a supervised learning task, each passive client $k$ maintains a dataset that consists of features $\mathcal{D}_k = \left\{ \boldsymbol{x}^k_i \right\}_{i=1}^{I} = \left\{ \left\{ x_{i,j}^k \right\}_{j \in \mathcal{J}^k} \right\}_{i=1}^{I}
$ characterizing $I$ total samples, and $j\in \mathcal{J}_k$ defines the indexes of the features housed at client $k$. Moreover, each passive client has a local model $f_{\theta_k} :\mathbb{R}^{|\mathcal{J}_k|}\rightarrow\mathcal{Z}_k$, parametrized by $\theta_k$ that takes users' $k$ allocated features characterizing each sample $i$ as an input and outputs an embedding $\boldsymbol{z}^k_i=f_{\theta^k}(\{x_{i,j}^k\}_{j\in\mathcal{J}_k})\in \mathcal{Z}_k$. Note that, in the prototypical VFL setting $\bigcup^K_{k=1}\mathcal{J}_k=\mathcal{J} $, where $\mathcal{J}$ denotes the set of features characterizing the task learned $\mathcal{T}:\mathcal{J}\rightarrow \mathcal{Y}$. However, there are no constraints regarding the intersection of feature sets between clients, denoted as $\mathcal{J}_k\bigcap\mathcal{J}_l$ for any distinct clients $k$ and $l$, suggesting that features may overlap across different clients. The \textit{active} client houses the samples ground truth labels $\textbf{Y} = \{\textbf{y}_i\}^I_{i=1}$, and is endued with a global model $M_{\theta_s} : \prod_{k=1}^K\mathcal{Z}_k\rightarrow \mathcal{Y}$, parameterized by $\theta_s$ that takes the embeddings $\textbf{Z}_i=(\boldsymbol{z}_i^1,\cdots,\boldsymbol{z}_i^k)$ of the passive clients as an input and outputs a prediction $\hat{\textbf{y}}_i\in\mathcal{Y}$.
We denote by $\hat{\textbf{y}}_i = M_{\theta_s}(\textbf{Z}_i)$ the prediction made at the server for sample $i$.
The VFL objective is to train a distributed model parameterized by  $\phi=\left(\{\theta_k\}^K_{k=1},\theta_s\right)$ that is able to minimize the empirical risk across all samples. Formally, this is given : 
\begin{equation*}
 R(\phi) =    \min_{\phi}\frac{1}{I}\sum^I_{i=1}\ell\left(\hat{\textbf{y}}_i,\textbf{y}_i\right),
\end{equation*}
where $\ell: \mathcal{Y}\times\mathcal{Y}\rightarrow\mathbb{R}$ denotes a loss function that computes the deviation between the prediction $\hat{\textbf{y}}$ and the ground truth label $\textbf{y}$. The training of the model happens across multiple communication rounds, similar to HFL. Particularly, at each round, the server receives a batch of embeddings from the different passive clients corresponding to a batch of samples. The server propagates those embeddings to predict the samples' labels. The error is computed and the gradient with respect to the server model is back-propagated. Furthermore, the gradient of the loss with respect to the client models is computed and sent to the clients. Those gradients are then back-propagated by the clients over their local models weights. This training process iterates until the model converges. During inference, the trained local and the server models are used to predict the labels corresponding to the data samples  at the passive clients.

\section{Related Work}
Conventional VFL methods typically assume ideal communication conditions and consistent availability of passive clients throughout both training and inference. However, this assumption often breaks down in practical VFL deployments. To address this, newer techniques have been proposed that explicitly consider client reliability. In earlier work, \cite{dropout} tackled the degradation in model performance caused by client dropout during inference by simulating dropout events during training, thus allowing the VFL model to adapt to such disruptions. While effective, this strategy depends on prior knowledge of each client's dropout probability during inference—information that is rarely available during the early stages of training. Building on this, \cite{unreliableserver} extends the approach to handle communication failures in cross-device VFL, while also accounting for potential server unreliability. However, both methods presume the presence of all clients during training, effectively requiring fully-observed training data—an assumption that often does not hold in practice. Additionally, these solutions are narrow in scope and may not be applicable to more controlled, centralized environments. Specifically, they overlook scenarios where feature assignment is influenced by factors beyond the data collection capabilities of individual passive clients. In such contexts, clients can be assigned feature partitions from a centrally stored feature pool. This is especially relevant in the 5G CN, where a rich set of features can be collected from various network functions  (NFs) and stored for downstream processing at the CN \cite{b1}.
\section{System Model}
We consider a CN architecture which adopts a 5G service based architecture (SBA) design principles  that are enriched by specialized AI NFs. These NFs are implemented through NWDAFs, with each NWDAF containing its own MTLF and AnLF. The proposed CN architecture  alligns structurally with the CN model presented by \cite{jeon2022nwdaf}, featuring a structure with a centralized NWDAF and multiple NF-associated leaf NWDAFs. However, their fully-fledged design may introduce unnecessary redundancy. As an alternative, a more streamlined, less redundant approach is considered herein, where $N$  NWDAFs (NF-independent) and a NWDAF server are positioned in close proximity to critical NFs. The distinctive characteristic of FL deployment in the CN setting is data management flexibility. Unlike edge environments, the CN allows network data to be collected, stored, and distributed among NF under centralized coordination, leveraging full inter-NF connectivity, while introducing latency and signaling overhead. The data are gathered, managed, and stored by the Data Collection Coordination Function \cite{b2} and centralized databases managed by the Analytics Data Repository Function \cite{b3}. The data typically consists of network related measurements such as access network, transport network and application level measurements. Access network measurements encompass physical layer parameters, while application layer measurements incorporate quality of experience (QoE) metrics for various services. Additionally, data that users have explicitly authorized for processing (like location and mobility patterns) can be part of the measurements collected at NWDAFs. \textit{For notation simplicity,  we will hereafter refer to the passive NWDAFs selected for training as clients and the  NWDAF server simply as server.} 

The $N$ NWDAFs at the CN are characterized by a reliability defined by a probability $\boldsymbol{p} = \{p_1, \cdots, p_N\}$, denoting the probability that each client successfully completes its assigned task during a communication round in a time interval defined by the server. Dropout may occur when computational demands exceed the clients' resource threshold, preventing timely task completion within the designated communication interval. The selection of clients for VFL is influenced by the number of features and the complexity of the task. Allocating a large number of features to a limited number of clients typically necessitates larger models and increased memory usage, as the input dimensions of the local models and memory requirements scale with the number of assigned features.

Consequently, upon receiving a VFL request and during the initialization phase, the server determines the optimal number of clients to be selected, taking into account their computational and memory limitations, reliability, as well as their associated operational overhead. For simplicity, we assume that the server schedules $K \leq N$ clients and that their reliability is computed locally. Moreover, we assume that the server is fully reliable. However, other works have also considered scenarios where the server is unreliable \cite{unreliableserver}.

The VFL communication round is assumed to have a fixed time duration in which the embeddings should be computed and sent to the server. This duration is chosen by the server and communicated to the clients. Training operations specifically introduce substantial overhead due to their need to simultaneously maintain model parameters, store multiple data batches, and track gradient information. They also require approximately double the computational work, performing not just forward passes but also backward propagation and weight updates. When datasets aren't stored locally, the overhead is further amplified by data retrieval latency from other NF. These factors combined can potentially overwhelm a client, causing it to drop out.

Due to space limitations, we consider that the probability of such dropouts, represented by $\boldsymbol{p}$, encompasses all events that prevent a client from successfully completing its task within the given communication round. In the event that a client drops out during a communication round, it fails to compute or transmit its' embedding to the server. Under such circumstances, the server implements a baseline value as a substitute for the missing client embeddings. In our case, the embeddings are replaced with a $\vec{0}$ vector. Furthermore, in our model, we assume that a fixed communication budget is set, within which the embedding dimensions of the local models are determined.

\section{Proposed Algorithm}

\noindent Traditional VFL approaches assume uniform contributions from all clients, overlooking the variability in the value of insights contributed by individual clients based on their assigned or locally collected features. We capitalize on the controlled data access within the CN and propose an algorithm that leverages this flexibility. The proposed method enables the CN to manage feature analysis, distribution, and assignment while tailoring local clients models based on their reliability. Below, we outline the key steps of the algorithm, highlighting its workflow during model initialization, training and inference.

\subsection{Initialization of Training}

The training process begins when the network receives a training request. At this stage, the  server provides the available NWDAFs with detailed training specifications, which may include parameters such as the training duration, communication round length, and other relevant configurations.

Subsequently, the server prompts each  NWDAF to forecast their reliability for the anticipated training period. Each NWDAF estimates its reliability based on its current state, system conditions, and any relevant historical or projected performance data. This forecasted reliability information is then communicated back to the server to enable client selection and the subsequent model distribution and feature allocation strategies. The server then selects a set of \( K \)  clients based on their reliability, computational capabilities and operational overhead.
\begin{algorithm}[h!]
\small
\caption{Initialization of Training}
\label{alg:init}
\begin{algorithmic}[1]
  \State Network receives a training request.
  \State Server coordinates the training process. Communicate training specifications (e.g., training duration, communication round time) to available NWDAFs.
  \State Request each NWDAF  to forecast reliability over the training duration.
  \State Each NWDAF forecasts reliability based on current state, system conditions, and historical data.
  \State NWDAFs transmit their forecasted reliability measures back to the server.
  \State  Server selects $K$ clients based on reliability, computational capacity and overhead.
\end{algorithmic}
\end{algorithm}
\subsection{Feature Association and Model Definition}

After collecting the reliability measures from the NWDAFs and selecting the $K$  clients, the  server initiates a feature association and model definition phase. To begin with, an ANLF is instructed to perform feature alignment, preprocessing, and importance analysis on the data samples. Features from these centralized database samples can be analyzed through various techniques, including SHAPLEY value calculation \cite{b4}, mutual information assessment, or Decision Tree models to evaluate each feature's importance to the target learning objective. The analysis results are subsequently transmitted back to the  server.
\begin{algorithm}[h!]
\small
\caption{Feature Association and Model Definition}
\label{alg:feature}
\begin{algorithmic}[1]
  \State Server instructs ANLF to align, preprocess, and analyze feature importance (e.g., SHAPLEY values, mutual information, etc..).
  \State Associate features to  clients local models in proportion to their reliability probabilities.
  \State Define local model input and output dimensions based on assigned features and client reliability, under communication budget constraints.
  \State Construct server model architecture based on aggregated embedding dimensions.
  \State Distribute local models and assigned features identifiers to each client.
  \State Clients retrieve corresponding feature-aligned data samples from the central database.
\end{algorithmic}
\end{algorithm}
Feature association to the selected clients local models is subsequently carried out in proportion to the their reliability. In this scheme, clients with higher reliability probability are assigned features with greater importance. Consequently, the input and output dimensionality (i.e. the embeddings dimensions) of each model is determined based on the assigned features. The embeddings dimensionality may scale proportionally with both the number of features and the client's reliability probability, while ensuring compliance with a predefined communication budget. The number and type of layers in each local model can be also defined and adjusted according to the computational and memory constraints of their respective client.

 The individual models, along with their associated features, are then distributed to their respective clients. Highly reliable clients are assigned models associated with highly important features. To facilitate model training, each client requests the aligned data samples corresponding to its assigned features from the central database.

\subsection{Training}
Each client \( k \) is considered available during any given communication round with a probability \( p_k \). At each round, available clients perform forward passes through their respective local models using a batch of data samples. Those clients generate intermediate embeddings, which are subsequently transmitted to the central server. In contrast, unavailable clients embeddings are set to a $\vec{\boldsymbol{0}}$ vector by the server.

Upon receiving the embeddings, the server processes them through its model to generate predicted labels. The prediction error is then quantified using a loss function, and its' gradient is back-propagated to update the server-side model parameters. Additionally, the gradients corresponding to the local models embeddings are sent back to the available clients, enabling them to update their respective model weights. Unavailable clients during a communication round do not receive updates and therefore their local model is left unchanged. This iterative training process continues until the model converges or a predefined convergence criterion is met. The clients and server models are then stored; the local models are stored alongside a measure of their importance, determined by the importance of the features calculated in the previous step.
\begin{algorithm}[h!]
\small
\caption{Training Phase}
\label{alg:train}
\begin{algorithmic}[1]
  \While{convergence criterion not met}
    \For{each communication round}
      \State Available clients (each w.p $p_k$) perform forward pass on local data and send embeddings to server.
      \State Unavailable clients contribute no embeddings.
      \State Server aggregates embeddings and performs forward pass to generate predictions.
      \State Compute loss and back-propagate gradients to update server model.
      \State Transmit corresponding gradients to available clients for local model updates.
    \EndFor
  \EndWhile
  \State Store models; associate local models with their importance (based on feature significance) and complexity (based on memory requirements).
\end{algorithmic}
\end{algorithm}

\subsection{Inference}

When an inference request is initiated, the server first retrieves the forecasted reliability measures from the available NWDAFs. Based on these forecasts, the server selects the most reliable subset of $K$ NWDAFs that meet the necessary computational and memory capacity requirements by the stored models to participate in the inference task, ensuring compatibility with the complexity of the  stored models.

The selected clients are then assigned the trained local models based on their reliability measures, whereby higher reliability clients are associated with models of greater relative importance. Similar to the training phase, each client \( k \) is considered available during any given communication round with a probability \( p_k \). Moreover, the routing and alignment of the test datasets follows a similar approach to that defined during the training phase. Available clients during an inference round perform forward passes through their local models to generate the test embeddings, which are then transmitted to the server. Unavailable clients embeddings are set to $\vec{\boldsymbol{0}}$ by the server. The received embeddings are concatenated and processed by the server through its' model to predict the labels. 

\begin{algorithm}[H]
\small
\caption{Inference Phase}
\label{alg:infer}
\begin{algorithmic}[1]
  \State Retrieve forecasted reliability measures from available NWDAFs.
  \State Select top $K$ NWDAFs that meet reliability, computational, and memory requirements relative to model complexity.
  \State Assign trained local models to selected clients based on reliability, with more important models allocated to more reliable clients.
  \For{each inference round}
    \State Available clients (each w.p $p_k$)  perform forward pass and transmit embeddings to server.
    \State Unavailable clients contribute no embeddings.
    \State Server processes aggregated embeddings to produce final predictions.
  \EndFor
\end{algorithmic}
\end{algorithm}
\section{Simulation Results}\label{AA}
We evaluate our algorithm in comparison to the SoTA VFL implementation under client unreliability scenarios, as described in \cite{dropout}. In this setting, features are randomly distributed across  clients without considering their reliability, and client dropout naturally occurs due to their practical unreliability during both training and testing. Note that the baseline performance represents an upper bound on the performance of the algorithm described in \cite{dropout}, given that the reliability of clients remains consistent throughout both training and testing.
\subsection{Dataset}
The dataset used in this work is collected as part of the DASHing Factory project at Mitsubishi Electric R\&D Centre, focusing on Dynamic Adaptive Streaming over HTTP (DASH) within the context of a 5G factory automation scenario. The dataset is generated using ns-3 simulator with  5G-LENA and dash modules (responsible respectively of 5g NR and DASH streaming application). The simulation platform outputs a set of trace files covering various levels of the simulated scenario (mac, rlc, etc). Raw traces are then aggregated, cleaned and exported as a CSV file. The dataset is made of 69 features, and a QoE indicator serving as the label we wish to predict.  The simulation code is built upon FLOWER framework\cite{beutel2020flower}.  The dataset, simulation code, and training configuration can be accessed at  \href{https://github.com/merce-fra/Reliable-VFL-in-5G-Core-NW-Arch}{{\emph{https://github.com/merce-fra/Reliable-VFL-in-5G-Core-NW-Arch}}}.

\subsection{Simulation Model}
We choose $K=4$ NWDAFs to partake the role of clients and explore four different scenarios. In the first scenario, during each simulation run, each client success probability $p_k$ is drawn from a Beta(8,2) distribution, with a mean $\mu = 0.8$ indicating highly reliable setting. In the second scenario, the success probabilities are drawn from a Beta(10,6) distribution, with a mean  $\mu = 0.625$ signifying a moderately reliable setting.  During each simulation, the drawn success probabilities are  kept fixed throughout the training and test phases, and the availability of the clients during each communication round is determined according to a bernoulli distribution $a_k\sim \textnormal{Bern}(p_k)$. The total embeddings dimension is set to 48. Each scenario's performance is averaged over 10 simulation runs, with clients success probabilities sampled anew in each to reflect their reliability profiles during both training and testing. For simplicity, we allow the server to calculate the feature importances using a basic scikit Decision Tree Regressor\cite{scikit}. 

To evaluate the performance of our models under the proposed method against the SoTA baseline, we first determine each client success probability at the start of each simulation using via Beta distribution sampling. The clients are then tagged with $\{1,2,4,8\}$ in ascending order of reliability. These identifiers help examine model performance across varying  reliability levels. As previously mentioned, our framework assigns more important features and larger embedding dimensions to the most reliable clients. This is accomplished by assigning features to each client model so that the total importance of the associated features increases linearly with the client's reliability. The same principle applies to the embedding dimensions of each model, which scale linearly with the client's reliability. During training, we opt to minimize the Huber loss between the ground truth QoE label and the predicted QoE. We also continually store the best-performing server and client models in terms of its test performance, computed during each round, given the clients availability. Once training concludes, we revert to the top performing models to compare the two algorithms. Fig. 1 and 3 display the histograms of the average weighted test loss for the best model from scenario 1 (Beta(8,2)) and scenario 2 (Beta(10,6)) under different client participation patterns (averaged over 10 simulations runs). The x-axis, through the 15 IDs, designates these patterns. For instance, with given tags, an x-value (i.e., an ID) of 9 refers to the case where the most reliable client (tag 8) and the least reliable client (tag 1) partake in the test phase, as their tags' sum results in 9. Thus, $\textnormal{ID} = \sum^K_{k=1}a_k \textnormal{tag}_k$, where $a_k\sim \textnormal{Bern}(p_k)$ signifies whether client $k$ is available during an inference round. Sampled scenarios with no participating clients (i.e., ID = 0) are disregarded. Models are trained and their top performers (average test loss measured over test data and different availability patterns) are tested over 1000 rounds on the test dataset. For each scenario, the test loss is firstly weighted by the frequency of different IDs appearing in each simulation and then averaged over the 10 iterations. Accordingly, for a given ID pattern $m$, the weighted average loss $\ell(m)$ across all simulation runs can be expressed as:

\begin{equation*}
\ell_{\textnormal{test}}(\textnormal{ID}=m) = \frac{1}{N_{\text{sim}}} \sum_{n=1}^{N_{\text{sim}}} \ell_{\textnormal{test}}(m|n) \cdot f(m|n),
\end{equation*}

\noindent where $N_{\text{sim}}$ denotes the number of simulation runs, $\ell_{\textnormal{test}}(m|n)$ represents the average test loss (over the test dataset) observed for ID pattern $m$ in simulation run $n$, and $f(m|n)$ is the frequency of occurrence of pattern $m$ during the test phase of run $n$. On the other hand, Fig. 2 and 4 show the difference in weighted test loss between both algorithms in scenario 1 and 2, respectively. A negative difference at a given ID implies that our proposed method results in a lower average weighted loss compared to the baseline algorithm.
\begin{figure}[h]
    \centering
    \includegraphics[width=0.46\textwidth]{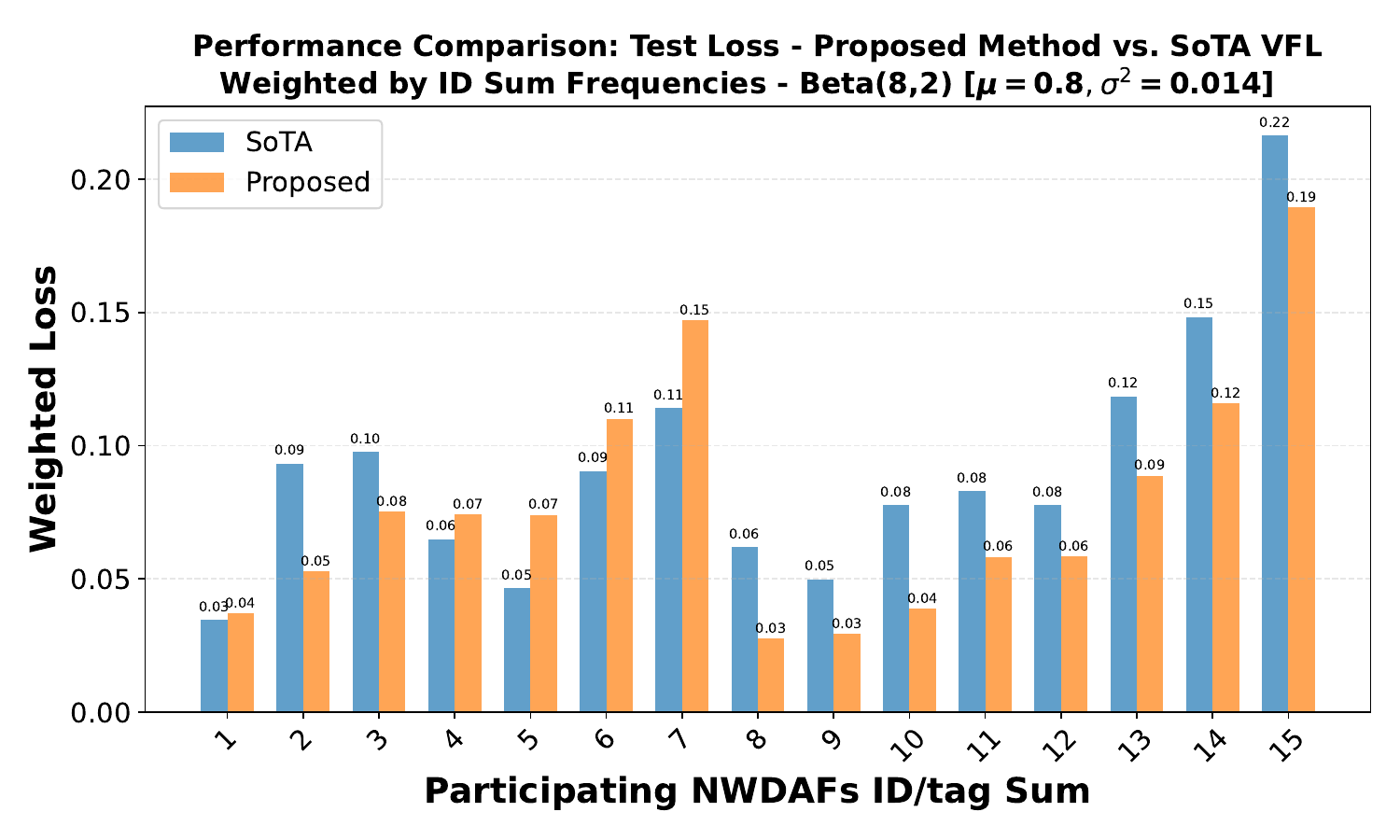} 
    \caption{\small \centering Weighted average test loss in scenario 1 ($p_k\sim \text{Beta(8,2)}$, averaged over 10 runs). x-axis represents the availability patterns based on assigned tags.}
    \label{fig:test_loss_1}
\end{figure}
\begin{figure}[h]
    \centering
    \includegraphics[width=0.46\textwidth]{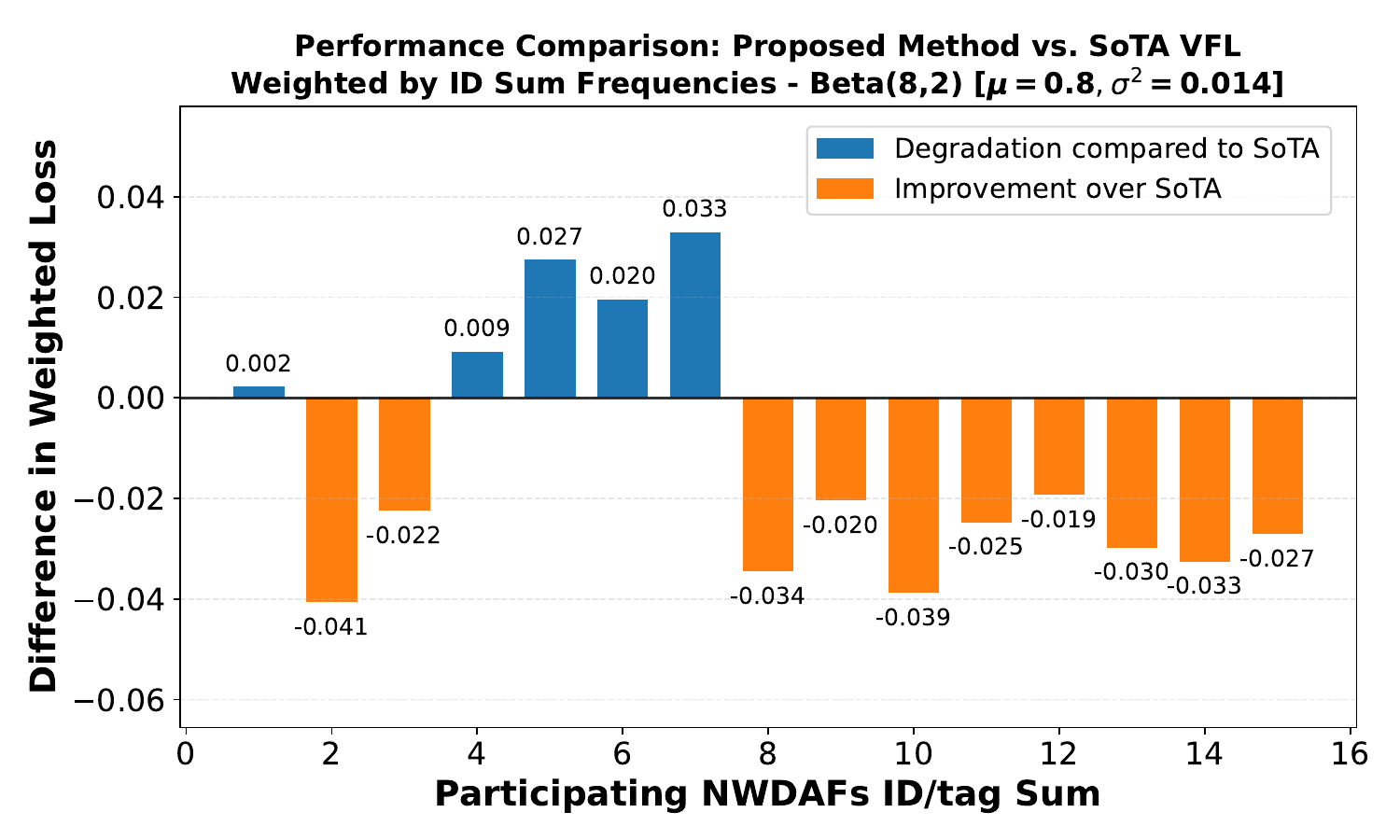} 
 \caption{\small \centering  Weighted test loss difference across the different ID patterns under scenario 1.}
    \label{fig:weighted_diff_1}
\end{figure}
\begin{figure}[h]
    \centering
    \includegraphics[width=0.46\textwidth]{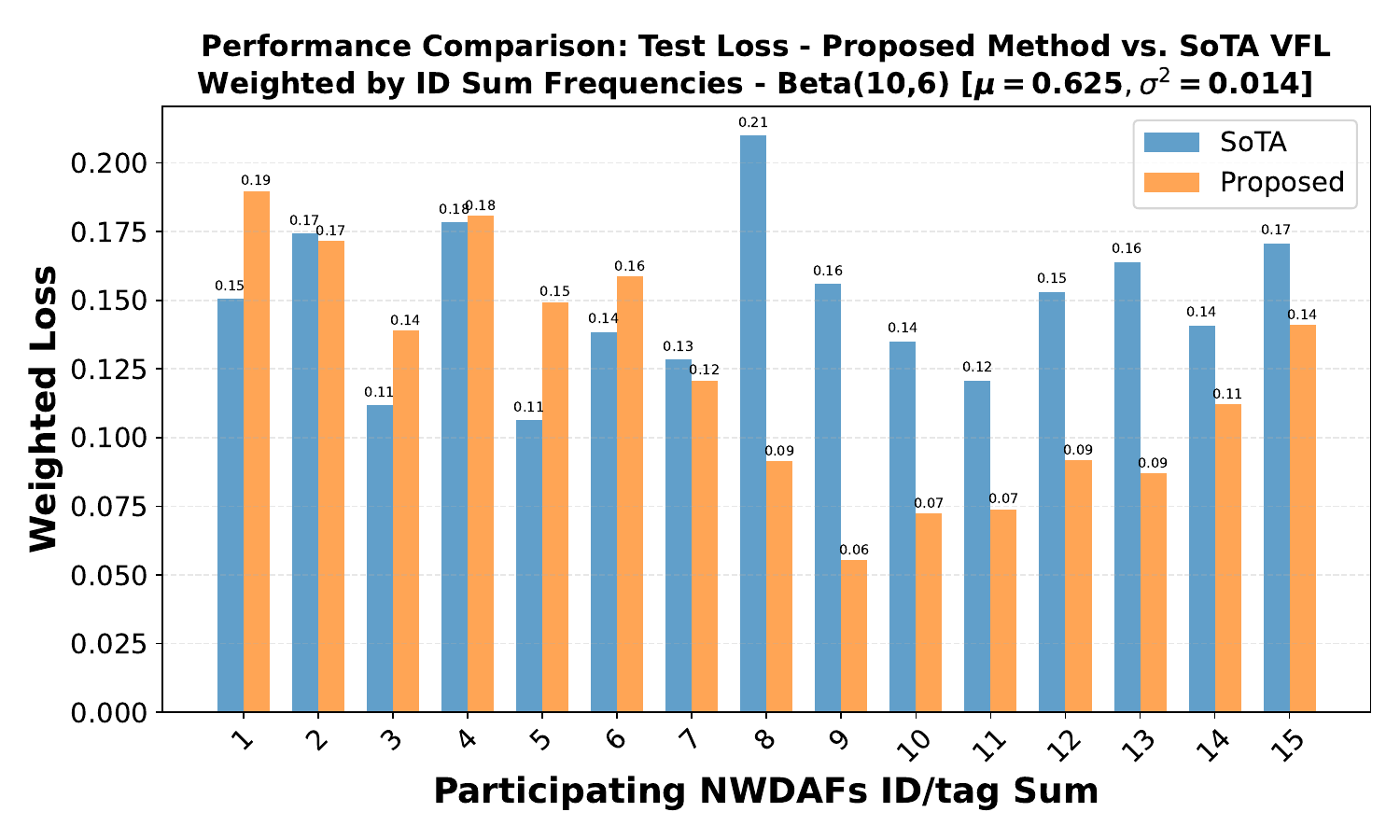} 
    \caption{\small \centering  Weighted average test loss in scenario 2 ($p_k\sim \text{Beta(10,6)}$, averaged over 10 runs). x-axis represents the availability patterns based on assigned tags.}
    \label{fig:test_loss_2}
\end{figure}
\begin{figure}[h]
    \centering
    \includegraphics[width=0.46\textwidth]{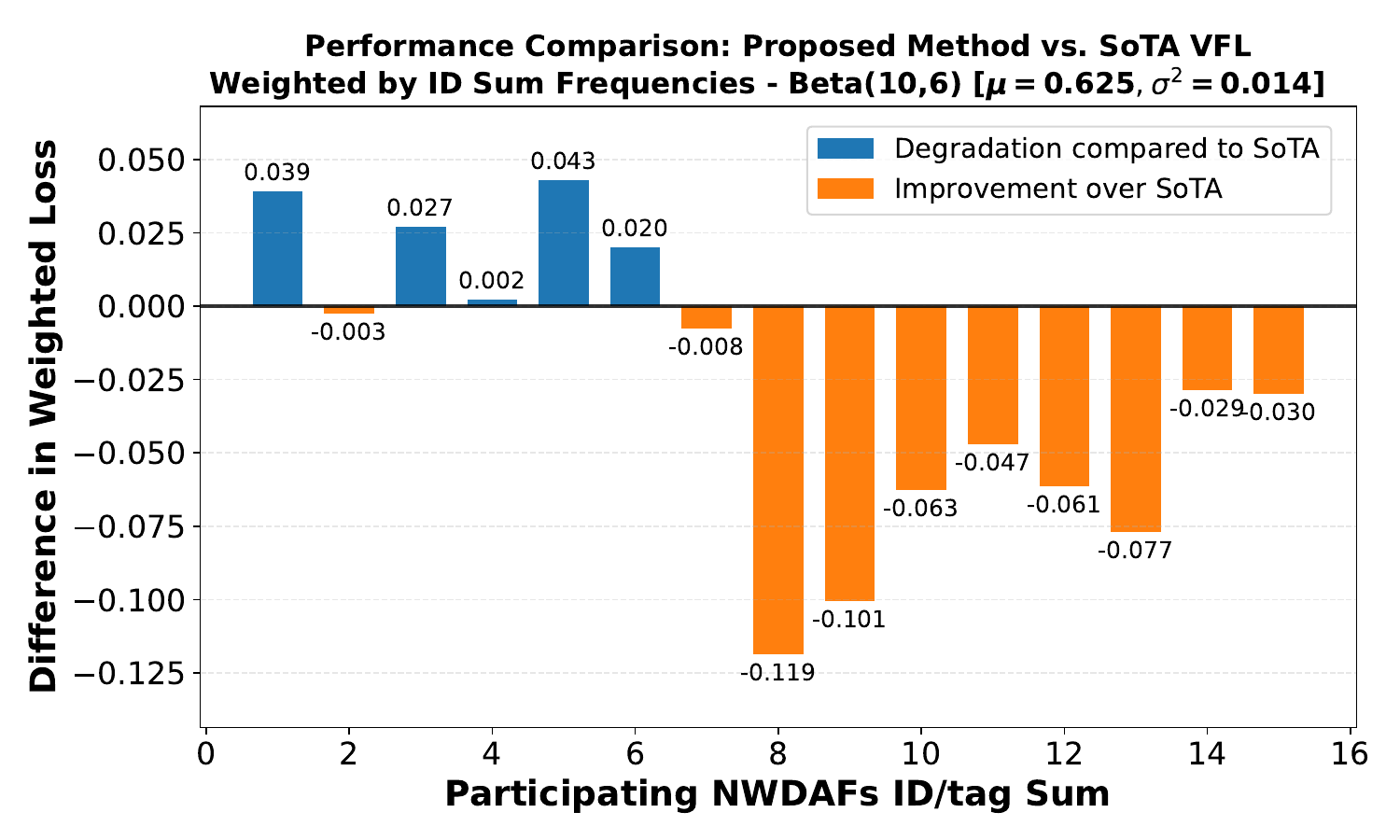} 
     \caption{\small \centering  Weighted test loss difference across the different ID patterns under scenario 2.}
    \label{fig:weighted_diff_2}
\end{figure}
\subsection{Results and Analysis}
Fig. 1-3 and 2-4 illustrate the average weighted loss and their difference of both approaches under scenario one and two respectively. Our proposed method achieves a lower test loss compared to the VFL baseline in both scenarios, with this improvement clearly reflected in the moderately reliable setting (Scenario 2). For scenario 1, characterized by highly reliable clients ($p_k \sim \text{Beta}(8,2)$), the performance advantage is relatively lower than the moderately reliable setting. Nevertheless, in this setting, our proposed algorithm still delivers approximately 14.5\% lower test loss on average across all client availability patterns when compared to the baseline approach. The percentage is calculated based on the following expression:
$$\sum^{15}_{m=1} \frac{|\ell^{\,\textnormal{proposed}}_{\textnormal{test}}(\textnormal{ID}=m) - \ell^{\,\textnormal{SoTA}}_{\textnormal{test}}(\textnormal{ID}=m)|}{\sum^{15}_{m=1}\ell^{\,\textnormal{SoTA}}_{\textnormal{test}}(\textnormal{ID}=m)}.$$
The performance gap is more evident in scenario 2, which features lower client reliability ($p_k \sim \text{Beta}(10,6)$). Specifically, our algorithm achieves  $\sim$18\% reduction in average test loss compared to the baseline in this scenario. It’s important to note that the improvement achieved by our proposed algorithm mainly stems from scenarios where the most reliable clients are involved in training, as evident in the Fig.4 —especially when ID$\ge$8, indicating that the most reliable client, associated with the most significant features, is available during testing.

\section{Conclusion}

In this work, we addressed the challenge of client reliability in VFL within 5G CNs. We proposed a novel feature distribution algorithm that  assigns important features to clients based on their reliability metrics, ensuring that critical features are allocated to the most reliable NWDAFs. Evaluation results demonstrate that our approach improves model performance under diverse client availability scenarios, with the most substantial gains observed in moderately reliable settings. Our approach leverages the centralized data orchestration capabilities of 5G CNs to implement an adaptive model design strategy that scales the clients' model input and embedding dimensions based on both feature importance and client reliability. The results underscore the importance of considering client reliability in VFL deployments, particularly in 5G CN environments where NWDAFs may experience varying levels of availability due to resource constraints and load overhead.

\vspace{12pt}

\end{document}